\documentclass{article}

\usepackage{arxiv}
\usepackage{amsmath}
\usepackage[utf8]{inputenc} % allow utf-8 input
\usepackage[T1]{fontenc}    % use 8-bit T1 fonts
\usepackage{hyperref}       % hyperlinks
\usepackage{url}            % simple URL typesetting
\usepackage{booktabs}       % professional-quality tables
\usepackage{amsfonts}       % blackboard math symbols
\usepackage{nicefrac}       % compact symbols for 1/2, etc.
\usepackage{microtype}      % microtypography
\usepackage{cleveref}       % smart cross-referencing
\usepackage{lipsum}         % Can be removed after putting your text content

\usepackage{graphicx}
\usepackage{doi}
\usepackage{multirow} 
\title{High Dynamic Range Imaging Based on an Asymmetric Event-SVE Camera System}

\newif\ifuniqueAffiliation
\uniqueAffiliationtrue

\usepackage{authblk}

\author[1,2]{Pengju Sun,}
\author[1,2,*]{Banglei Guan}
\author[1,2,*]{Jing Tao}
\author[1,2]{Zhenbao Yu}
\author[1,2]{Xuanyu Bai}
\author[1,2,*]{Yang Shang}
\author[1,2]{Qifeng Yu}
\affil[1]{College of Aerospace Science and Engineering, National University of Defense Technology, Hunan 410073, China}
\affil[2]{Hunan Provincial Key Laboratory of Image Measurement and Vision Navigation, Hunan 410073, China}

\renewcommand{\shorttitle}{Applied Optics}

\hypersetup{
pdftitle={A template for the arxiv style},
pdfsubject={q-bio.NC, q-bio.QM},
pdfauthor={David S.~Hippocampus, Elias D.~Striatum},
pdfkeywords={First keyword, Second keyword, More},
}

\begin{document}
\maketitle
\begin{abstract}
	High dynamic range (HDR) imaging under extreme illumination remains challenging for conventional cameras due to overexposure. Event cameras provide microsecond temporal resolution and high dynamic range, while spatially varying exposure (SVE) sensors offer single-shot radiometric diversity.We present a hardware--algorithm co-designed HDR imaging system that tightly integrates an SVE micro-attenuation camera with an event sensor in an asymmetric dual-modality configuration. To handle non-coaxial geometry and heterogeneous optics,  we develop a two-stage cross-modal alignment framework that combines feature-guided coarse homography estimation with a multi-scale refinement module based on spatial pooling and frequency-domain filtering. On top of aligned representations, we develop a cross-modal HDR reconstruction network with convolutional fusion, mutual-information regularization, and a learnable fusion loss that adaptively balances intensity cues and event-derived structural constraints. Comprehensive experiments on both synthetic benchmarks and real captures demonstrate that the proposed system consistently improves highlight recovery, edge fidelity, and robustness compared with frame-only or event-only HDR pipelines. The results indicate that jointly optimizing optical design, cross-modal alignment, and computational fusion provides an effective foundation for reliable HDR perception in highly dynamic and radiometrically challenging environments.
\end{abstract}

\section{Introduction}
\label{sec:intro}

HDR imaging has become increasingly essential across modern computer vision, robotics, and scientific sensing systems, where environments often exhibit radiance variations that far exceed the limits of conventional cameras~\cite{gehrig2024low,HDRev2023learningcvpr,hasinoff2016burst,messikommer2022multi,tan2026optimal,Deng_OE_25hdr}.
Real-world scenes such as nighttime urban navigation, high-speed industrial inspection, autonomous driving through tunnels, and scientific experiments involving rapidly changing illumination routinely span tens of thousands of brightness levels~\cite{Jiang2024EventBasedLI,Guan2025IJCV,Huangao25}. In such conditions, standard frame-based cameras struggle to capture both bright and dark regions simultaneously, leading to saturation, loss of structural detail, and unreliable perception for downstream tasks~\cite{pami2024eventhdr,TPAMI_Guan,ijcv2021,liu2023low}. Multi-exposure bracketing and learned HDR reconstruction have greatly progressed. However, they still face the dual challenge of capturing a scene's full brightness range while suppressing motion blur and ghosting~\cite{messikommer2022multi}. In particular, they remain limited by the dynamic range and readout speed of standard image sensors.

To expand the radiometric range with minimal acquisition overhead, a rich body of work explores spatial and temporal exposure multiplexing. Multi-camera arrays acquire multiple exposures simultaneously using different cameras~\cite{Kronander2013AUF,9009997}, but inevitably introduce parallax and inter-camera photometric differences that require robust matching and alignment~\cite{Lee2017StereoMU}. Classical HDR pipelines fuse temporally bracketed exposures using inverse CRF–based merging or exposure fusion, which can still be prone to failure under extreme radiance spans~\cite{bernacki2020automatic,Sun_OE_25HDR}. Single-image HDR methods attempt to hallucinate missing detail from a single low dynamic range input, but they struggle to reliably recover saturated highlights~\cite{yuan2012automatic}. More recently, attention-inspired architectures have been employed to perform implicit alignment and artifact-aware fusion across frames~\cite{HDRECCV2022,cai2023retinexformer,liu2023holoco}, yet these methods are {fundamentally constrained by the limited dynamic range}.

Spatially varying exposure (SVE) cameras provide a promising alternative by implementing micro-filter patterns that yield a spatial mosaic of effective exposure ratios within a single readout~\cite{2003iccv,acmtog2017,TAO2025olt}. {Such designs reduce ghosting relative to temporal bracketing and provide a bracketed stack without extra capture time~\cite{ME2021}. Reconstruction} from mosaicked exposures entails demosaicking across large brightness disparities, which is error-prone in textureless regions and around high-contrast edges~\cite{IROS2022sve}. It is designed to integrate seamlessly with standard optics while delivering improved HDR performance compared to traditional cameras~\cite{IROS2022sve}.

Event cameras provide a complementary sensing modality. Instead of outputting frames at fixed intervals, they asynchronously report per-pixel brightness changes with microsecond latency and extremely large dynamic range~\cite{gallego2020event,LeiOE_event}. These properties directly {address saturation that} hinder conventional HDR pipelines and make event cameras attractive for HDR video reconstruction~\cite{7758089,Censi2014LowlatencyEV,9156346,HDRev2023learningcvpr}. However, event sensors do not directly {measure absolute intensity,} and they suffer from contrast-threshold bias, sensor noise, and the lack of a well-defined radiometric scale~\cite{afifi2021learning}. Event-only HDR reconstruction can therefore accumulate errors over time and often yields {images with inaccurate grayscale information}.

These complementary strengths and weaknesses naturally motivate cross-modal fusion of event streams and frames. A series of recent methods leverage events to deblur or enhance low-light frames, or to reconstruct HDR-like imagery under extreme illumination. ELIE composes underexposed frames with event streams via residual-level fusion and multi-level reconstruction losses, and introduces the LIE dataset, demonstrating substantial gains over frame-only methods~\cite{Jiang2024EventBasedLI}. Building on large-scale, precisely aligned real captures, EvLight proposes SNR-guided regional feature selection and a holistic-regional fusion branch, reporting consistent improvements on both real and synthetic datasets~\cite{liang2024towards}. In parallel, recurrent event-based HDR video reconstruction frameworks~\cite{pami2024eventhdr} have shown that event sequences can drive high-speed HDR reconstruction when combined with key-frame guidance.

\begin{figure*}[t]
\centering
\includegraphics[width=0.95\linewidth]{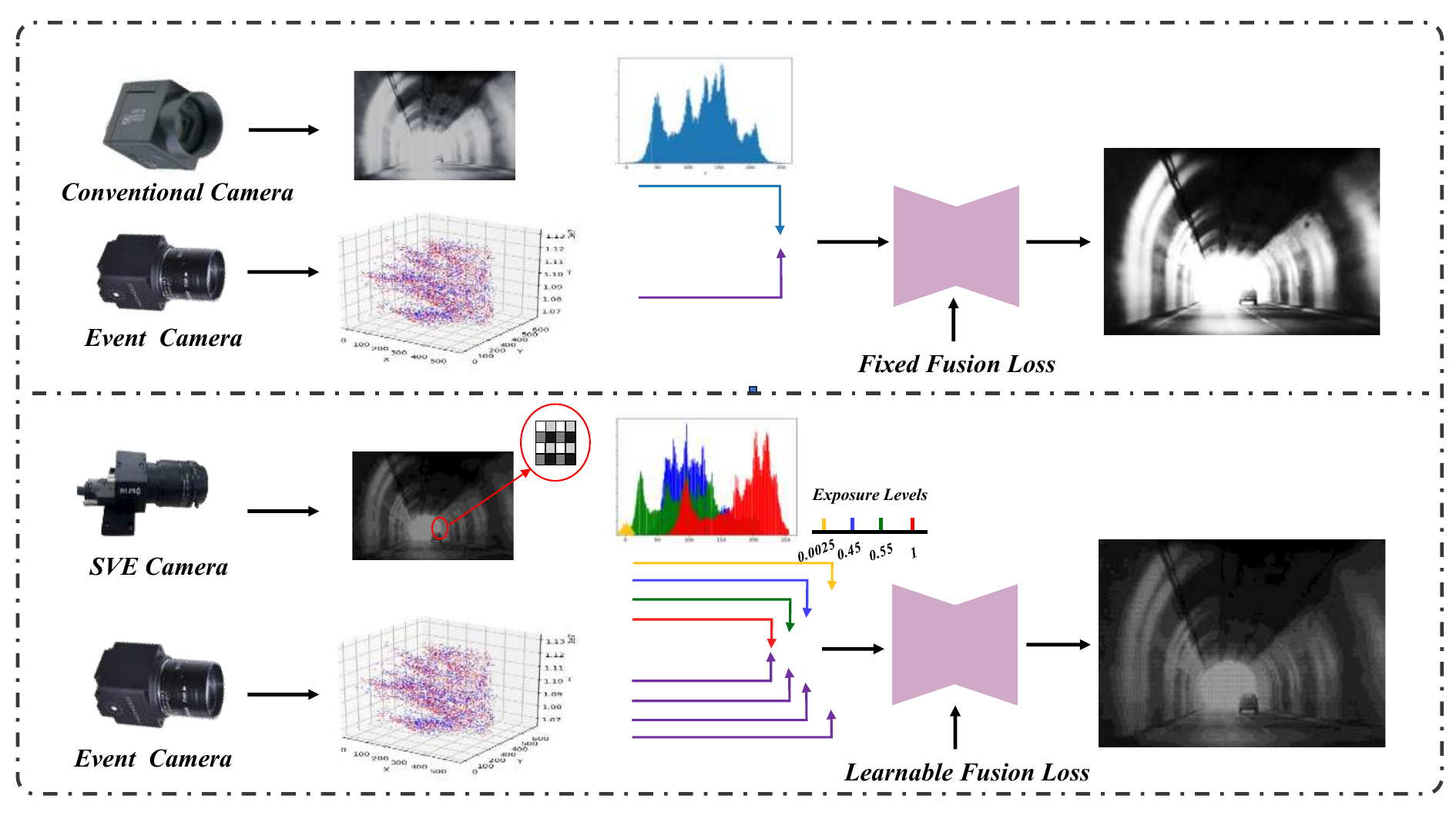}
\caption{Overview of event-assisted HDR reconstruction. {Top: a conventional event+frame HDR pipeline typically fuses an intensity frame with an event representation using fixed fusion loss}.
Bottom: {our asymmetric Event-SVE HDR system encodes multi-exposure radiometric measurements and performs calibration-guided coarse alignment followed by learnable refinement due to the non-coaxial setup.}
} 
\label{fig:teaser}
\end{figure*}

Despite these advances, existing fusion frameworks are typically designed for {conventional cameras} or beam-splitter rigs~\cite{Wu_OE_24_event_HDR}. {Many methods are limited by assumptions of coaxiality, small baselines, or homogeneous exposure, and do not sufficiently address the specific radiometric and geometric properties of micro-attenuation mosaics.} Moreover, most current methods rely on manually crafted fusion losses that obey fixed design guidelines~\cite{sun2022event,evlowlight2023iccv}. Such predefined terms may capture generic priors but lack the flexibility to adapt to different exposure patterns, scene dynamics, or downstream tasks. In particular, they provide limited capacity for learnable reweighting of event and frame contributions within HDR reconstruction.

In many practical systems, an event camera and a conventional camera are mounted non-coaxially  {with distinct optics, focal lengths, and fields of view}. {The heterogeneous imaging geometry, along with asynchronous sensing and varying sampling densities, poses significant challenges for correspondence, calibration, and fusion}~\cite{ijcv2021,Tulyakov21CVPR,tulyakov2022time}. While single-exposure frames offer restricted intensity information, their integration with event data enables new approaches to HDR reconstruction~\cite{cui2024color,HDRev2023learningcvpr}. However, a comprehensive solution that co-designs hardware and algorithms, explicitly handles non-coaxial geometry across heterogeneous sensors, and performs learnable fusion for HDR reconstruction remains underexplored.

To this end, we consider an asymmetric hybrid imaging system that combines an event camera with an SVE camera, {exploiting the complementarity between two emerging modalities.} Our computational imaging system implements a synchronized {multi-modality acquisition strategy}. The SVE sensor delivers spatially dense multi-exposure images at a moderate frame rate, {while the event camera captures high-temporal-resolution brightness changes.}  {Unlike coaxial systems, our two sensors have independent optical paths and thus require explicit cross-modal alignment, as illustrated in Fig.~\ref{fig:teaser}.}
The SVE mosaic provides spatial radiometric diversity (multi-exposure cues) without temporal bracketing, while events provide temporally dense edge/motion information that is robust to blur and partial saturation.
{To address this, we implement an alignment-and-fusion pipeline (as opposed to pixel-wise merging) that adapts to the radiometric heterogeneity of the SVE mosaic and the sparsity of event streams, leading to improved highlight recovery and texture fidelity.}

In summary, the main contributions of this work are as follows:
\begin{itemize}
\item An asymmetric Event–SVE HDR imaging prototype is built with independent optical paths, and a synchronized acquisition/processing pipeline is provided to pair each SVE exposure cycle with its corresponding event interval.
\item A two-stage cross-modal alignment framework is proposed for non-coaxial geometry, consisting of calibration-guided coarse homography rectification and a learnable multi-scale refinement module leveraging spatial pooling and frequency-domain convolution to {mitigate residual parallax.}
\item We design a dual-branch HDR reconstruction network with cross-modal feature fusion and mutual-information regularization, and introduce a learnable fusion loss that predicts pixel-wise modality weights to adaptively balance SVE radiometric cues and event-driven structural constraints.
\end{itemize}

The remainder of this paper is organized as follows. Section~\ref{sec:proposed_method} details the hardware setup, non-coaxial alignment, and the proposed fusion network architecture and loss design. Section~\ref{sec:experiments} presents {quantitative, qualitative experimental results, and ablation analyses.} Finally, Section~\ref{Conclusion} concludes the paper.

\section{Proposed Method}
\label{sec:proposed_method}

To obtain HDR images from an asymmetric Event-SVE camera system, it is crucial to fully exploit the complementary advantages of event streams and SVE frames. This involves accurate spatio-temporal alignment, effective cross-modal feature fusion, and robust HDR reconstruction under extreme illumination. In this section, we present a unified framework for HDR reconstruction that integrates a non-coaxial SVE-event imaging system, a cross-modal coarse-to-fine alignment module, and a dual-path fusion network with a learnable fusion loss.

\subsection{Method Overview}

We design a hybrid HDR imaging pipeline that explicitly models both sensor characteristics. The framework follows a dual-stage architecture. The raw SVE frames are first decomposed via spatial demultiplexing into exposure-separated intensity maps. These multi-exposure images are concatenated and processed by an encoder-decoder network to integrate spatially distributed exposure information and obtain a radiometrically consistent SVE feature representation. In parallel, the event stream is accumulated into voxel grids and passed through an event encoder. A cross-modal fine alignment module then establishes content-adaptive cross-modal correspondence in feature space, producing geometrically consistent representations for subsequent HDR reconstruction.

On top of this architecture, we introduce a learnable fusion loss that adaptively modulates the contributions of SVE intensity cues and event-derived structural signals, guided by the HDR reconstruction objective.

\subsection{Formulation of the SVE-Event Imaging System}
\label{subsec:hardware}

\begin{figure*}[t]
\centering
\includegraphics[width=0.95\linewidth]{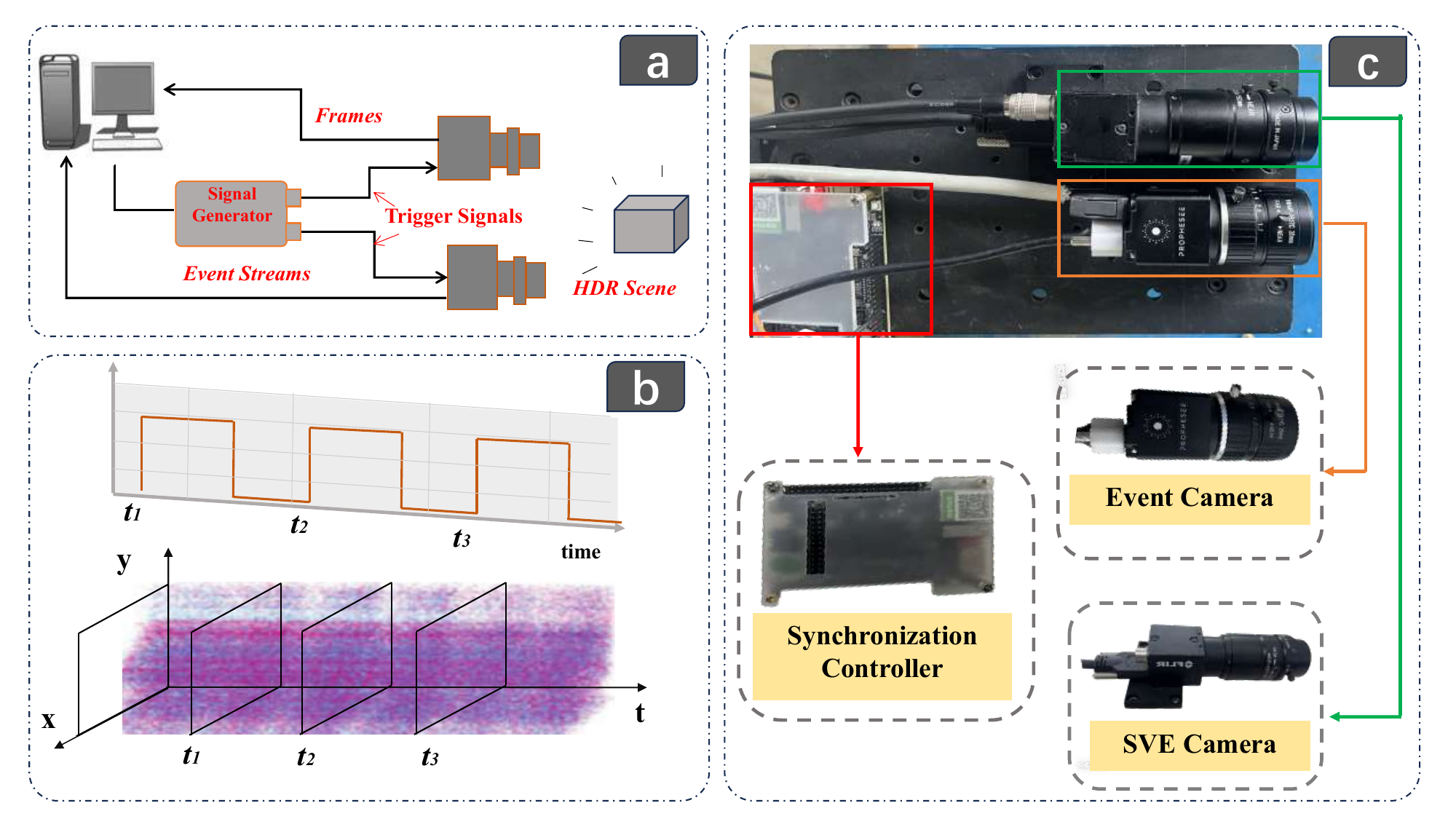}
\caption{Hybrid Event-SVE imaging platform. (a) Acquisition architecture in which a programmable trigger generator synchronizes the SVE camera’s exposure cycles with the event sensor’s asynchronous readout. (b) Temporal sampling characteristics showing discrete multi-exposure frame capture versus continuous event generation. (c) Hardware prototype with independent optical paths and a custom synchronization controller that provides a shared trigger reference for event-frame pairing.}
\label{fig:system}
\end{figure*}

We consider a non-coaxial hybrid imaging system that integrates an SVE camera with an event sensor, as shown in Fig.~\ref{fig:system}. {The SVE camera employs a $2\times2$ macro-pixel micro-attenuation mosaic to encode multiple radiometric measurements in a single shot. Let $E(\mathbf{x})$ denote the scene $\Phi(\mathbf{x})$ and $\tau_k$ the attenuation factor of the $k$-th sub-pixel ($k\in\{1,2,3,4\}$).} The sub-exposure measurement is modeled as
\begin{equation}
I_k(\mathbf{x}) = f\!\left(E(\mathbf{x})\,\tau_k\right),
\label{eq:sve}
\end{equation}
where $f(\cdot)$ is the camera response function. {In our hardware, $\{\tau_k\}$ are fixed after fabrication. In practice, the sensor outputs a single raw mosaic frame $M(\mathbf{u})$ in which each pixel location $\mathbf{u}$ belongs to one of the four sub-pixel types. Let $s(\mathbf{u})\in\{1,2,3,4\}$ denote the sub-pixel index at $\mathbf{u}$. The raw mosaic formation is
\begin{equation}
M(\mathbf{u}) \;=\; f\!\left(E(\mathbf{u})\,\tau_{s(\mathbf{u})}\,T_{\mathrm{exp}}\right),
\label{eq:sve_mosaic}
\end{equation}
where $T_{\mathrm{exp}}$ is the exposure time of one SVE cycle. This model shows that a single frame spatially interleaves four exposure levels. In our prototype, the four sub-pixels (top-left, top-right, bottom-left, bottom-right) are assigned 
(0.95, 0.45, 0.55, 0.005), respectively.
 We demultiplex the mosaic by grouping pixels with the same $s(\mathbf{u})$ to form four sub-images (one per $\tau_k$) for subsequent alignment and HDR reconstruction.}

Complementary to the SVE camera, the event sensor asynchronously records changes in logarithmic intensity. {An event $e_j=(x_j,y_j,t_j,p_j)$ with polarity $p_j\in\{+1,-1\}$ is triggered when the log-intensity change exceeds a contrast threshold $C$:
\begin{equation}
L(\mathbf{x},t)-L(\mathbf{x},t^-)=p\,C,\quad L(\mathbf{x},t)=\log I(\mathbf{x},t),
\label{eq:event}
\end{equation}
where $t^-$ is the most recent event time at pixel $\mathbf{x}$.} The event stream provides fine-grained temporal structural cues under rapid motion and extreme brightness changes, while the SVE camera supplies spatially dense multi-exposure sampling, forming a complementary sensing pair for HDR reconstruction.

{In our system, the event sensor is a Prophesee EVK4 with resolution $1280\times720$, and the SVE camera is our in-house prototype with native readout resolution $2048\times2448$ (effective $1024\times1224$ after $2\times2$ demultiplexing/processing). The two sensors share a hardware trigger at $60$\,Hz, where each pulse marks the start of one SVE exposure cycle; events are paired to frames by trigger index. Let $t_i$ denote the timestamp of the $i$-th trigger.  The camera-to-camera baseline is $50$\,mm, the SVE exposure time per cycle is $16$\,ms, and RAW readout is used to ensure consistent radiometric behavior.}

The event stream is ${E}=\{(x_j,y_j,t_j,p_j)\}_{j}$. {Within each $T_i$, we construct an accumulated event frame $\tilde{A}_i$ (single temporal bin, $B=1$) using a two-channel polarity encoding.} Each channel is normalized by the number of events in $T_i$, followed by clipping and linear rescaling to $[0,1]$. We adopt accumulated frames because they are lightweight compared with voxel grids while providing stable structural cues for cross-modal alignment and feature extraction. Specifically, {we compute modality-invariant correspondences between the undistorted SVE image and $\tilde{A}_i$ to estimate a global homography for coarse alignment, and then feed the warped $\tilde{A}_i$ into the event encoder for fine alignment and HDR fusion.}

This hardware-level synchronization scheme eliminates timestamp offsets. The shared trigger source guarantees deterministic timing with jitter below a single sensor clock cycle, providing substantially more reliable alignment than software-based timestamp association.

\begin{figure*}[t]
    \centering
    \includegraphics[width=0.95\linewidth]{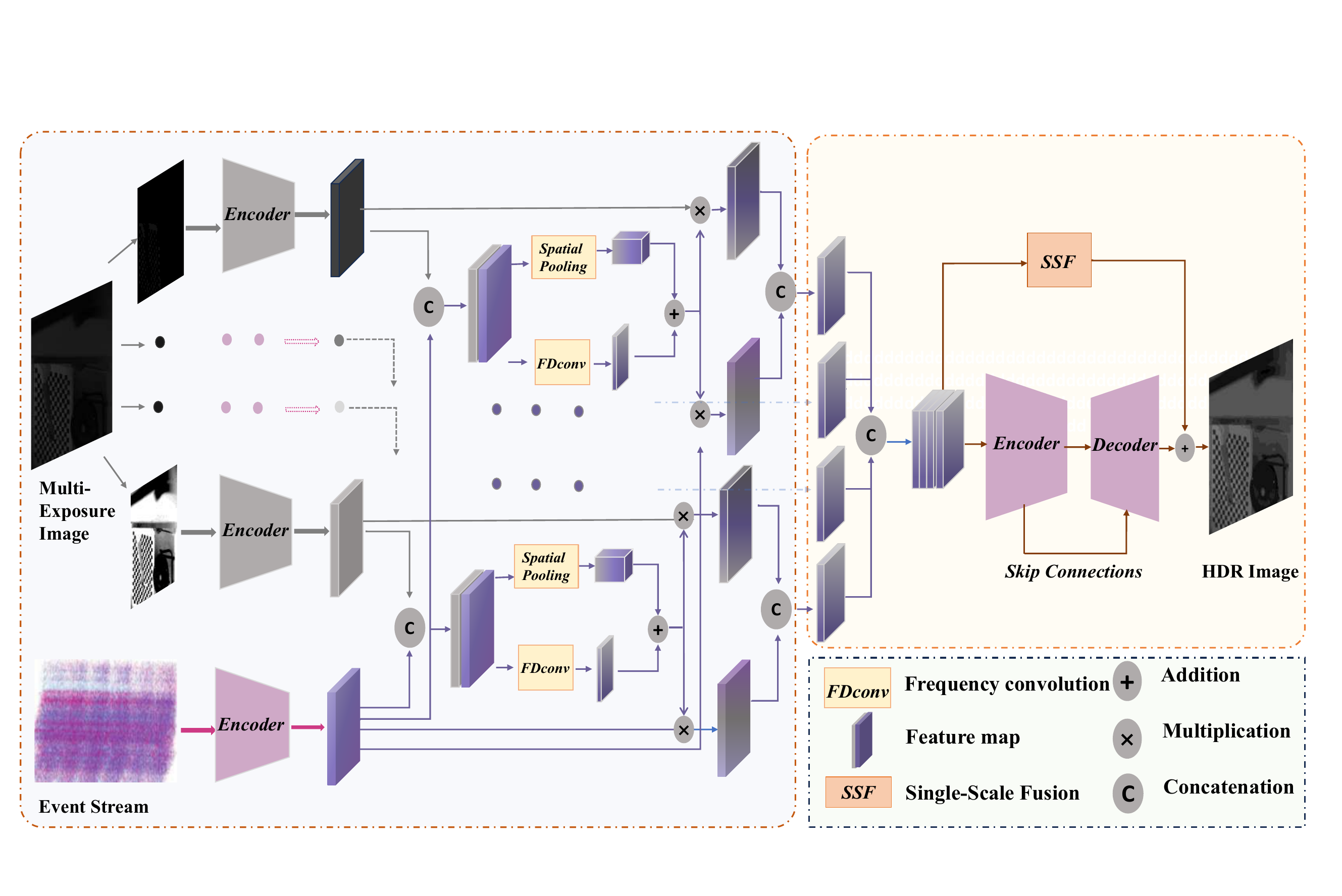}
\caption{Overview of the proposed cross-modal alignment and HDR fusion network. Multi-exposure SVE frames and event streams are encoded into multi-scale feature pyramids. At each pyramid level, the features are refined through spatial pooling and frequency-domain convolution to achieve cross-modal alignment. The aligned features are then aggregated and passed through an encoder–decoder sub-network to reconstruct the final HDR image.}
    \label{Overview}
\end{figure*}

\subsection{Cross-Modal HDR Reconstruction Network}

Our HDR reconstruction network adopts a {multi-branch} cross-modal architecture that jointly exploits multi-exposure intensity frames and event streams, as illustrated in Fig.~\ref{Overview}. The model begins with two modality-specific encoders that {extract spatial features from the SVE images and  event representation}. For the image branch, multi-exposure inputs are encoded into hierarchical feature pyramids, while the event branch processes the asynchronous event stream into a dense spatio-temporal embedding. At each scale, the corresponding features from the two branches are concatenated and subsequently enhanced through a combination of spatial pooling and frequency-domain convolution (FDConv), which selectively aggregates high-frequency structural cues and exposure-invariant information.

These fused multi-scale features are progressively propagated through a {UNet-like backbone}, where deeper layers capture global illumination context and shallow layers retain fine structural details. Skip connections are employed to preserve spatial fidelity and stabilize cross-modal information flow. The aggregated multi-level representation is then passed into a lightweight encoder-decoder reconstruction module equipped with a Single-Scale Fusion (SSF) block, which refines the fused features and produces a high-quality HDR image. This design effectively leverages the complementary properties of event-driven temporal dynamics and spatially varying exposures, enabling robust HDR reconstruction under challenging illumination conditions.

\subsubsection{Cross-Modal Coarse-to-Fine Alignment}
\label{subsec:alignment}

The asymmetric Event-SVE configuration employs independent optical paths, resulting in non-coaxial geometry, baseline-induced parallax, and modality-dependent appearance variations. Direct fusion without alignment leads to structural duplication and ghosting, especially around depth discontinuities. We therefore introduce a two-stage cross-modal alignment framework that combines coarse global rectification via feature-driven homography estimation and fine alignment through multi-scale cross-modal feature fusion.

We estimate a global homography $H^{\ast}$ for coarse alignment by minimizing the reprojection error of inlier matches:
\begin{equation}
H^{\ast} = \arg\min_{H}\sum_{i\in\mathcal{I}}
\left\|\pi\!\left(H\,\tilde{\mathbf{x}}^{\mathrm{SVE}}_{i}\right) - \mathbf{x}^{\mathrm{evt}}_{i}\right\|_2^{2},
\label{eq:homography}
\end{equation}
where $\tilde{\mathbf{x}}=[x,y,1]^{\top}$ denotes homogeneous coordinates and $\pi([u,v,w]^{\top})=[u/w,v/w]^{\top}$.
In practice, we extract tentative correspondences using an automatic detector-free matcher\cite{sun2021loftr} and robustly fit the homography via RANSAC;  manual point selection is only used as a fallback when automatic matching fails. Once estimated, the homography is kept fixed for the corresponding capture setup (or per-sequence if the mounting changes), and serves as a coarse initialization for the subsequent refinement stage.
{The coarse warp corrects global misalignment, while a multi-scale refinement module addresses residual local parallax from depth variations and fast motion.}

The coarse alignment produces a warped event representation
\begin{equation}
E^{\mathrm{warp}} = \mathcal{W}(E; \mathbf{H}^{\ast}),
\end{equation}
significantly reduces the global geometric discrepancies and provides a suitable initialization for fine alignment.

Despite coarse warping, {residual misalignment persists due to local depth variation, asynchronous sampling between events and frames. To correct these residual discrepancies, a learnable fine alignment module is required to achieve content-adaptive alignment.}

\subsubsection{Fine Alignment via Multi-Scale Cross-Modal Feature Fusion}

The fine alignment stage refines the spatial correspondence between modalities using a pyramid-based feature fusion architecture. Given the coarse-warped event representation $E_{\mathrm{warp}}$ and the input SVE image $I_{\mathrm{SVE}}$,
we first demultiplex the mosaic into four exposure images:
\begin{equation}
\{I^{(k)}\}_{k=1}^{4} = \mathcal{D}_{\mathrm{SVE}}\!\left(I_{\mathrm{SVE}}\right).
\label{eq:dsve}
\end{equation}
Modality-specific encoders produce multi-scale feature pyramids:
\begin{equation}
\{F^{\ell}_{\mathrm{SVE},k}\}_{\ell=1}^{L} = \mathrm{Enc_{SVE}}\!\left(I^{(k)}\right), \quad
\{F^{\ell}_{\mathrm{evt}}\}_{\ell=1}^{L} = \mathrm{Enc_{evt}}\!\left(E_{\mathrm{warp}}\right).
\label{eq:pyramids}
\end{equation}
We aggregate multi-exposure features by concatenation:
\begin{equation}
F^{\ell}_{\mathrm{SVE}} = \Psi\!\left(F^{\ell}_{\mathrm{SVE},1},F^{\ell}_{\mathrm{SVE},2},F^{\ell}_{\mathrm{SVE},3},F^{\ell}_{\mathrm{SVE},4}\right).
\label{eq:agg}
\end{equation}

Here, $\mathcal{D}_{\mathrm{SVE}}$ decomposes the mosaic into four exposure-specific images; 
$\mathrm{Enc}_{\mathrm{SVE},k}$ extracts multi-scale features from each exposure; and $\Psi$ aggregates 
them through concatenation. The event encoder $\mathrm{Enc}_{\mathrm{evt}}$ 
produces a parallel multi-scale representation.

\begin{figure}[t]
\centering
\includegraphics[width=0.8\linewidth]{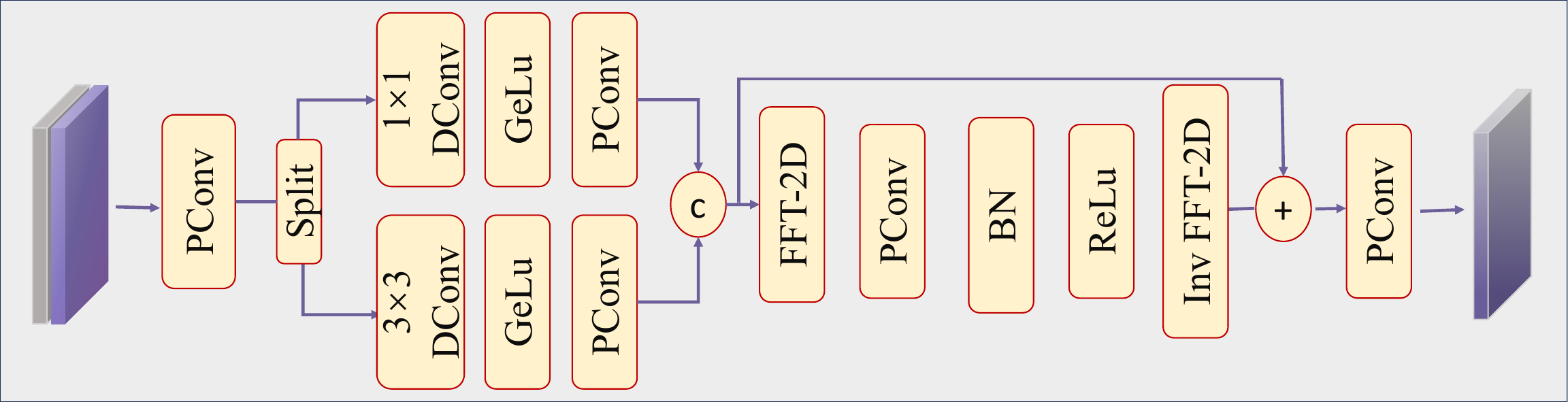}
\caption{{Structure diagram of the FDconv module. FDConv transforms features to the frequency domain, applies a learnable spectral response via element-wise complex multiplication, and transforms back.}
}
\label{FDConv}
\end{figure}

{To promote robust alignment, we employ two complementary operators at each scale including
 a spatial pooling module to stabilize local context and suppress modality-specific noise, and 
 a frequency-domain convolution, as shown in Fig.~\ref{FDConv}.}
Spatial pooling is introduced as a robust local aggregation operator to stabilize the matching cues under sparse or bursty events and modality-specific noise.
Frequency-domain convolution is adopted to explicitly separate structure-dominant components from radiometry-dominant components. {In practice, exposure changes and vignetting primarily perturb low-frequency components, whereas geometric misalignment is better constrained by high-frequency structures shared across modalities. Compared with purely spatial convolutions, FDConv provides a lightweight global-context operator and yields more stable gradients for alignment refinement when the two modalities have inconsistent photometric statistics.}

\begin{equation}
S_{\mathrm{SVE}}^{\ell} = \mathcal{P}(F_{\mathrm{SVE}}^{\ell}),
\qquad
S_{\mathrm{evt}}^{\ell} = \mathcal{P}(F_{\mathrm{evt}}^{\ell}),
\end{equation}
\begin{equation}
Z^{\ell}
= W_{c}^{\ell}
\big[
S_{\mathrm{SVE}}^{\ell}
\;\Vert\;
S_{\mathrm{evt}}^{\ell}
\big],
\end{equation}
We define FDConv as a spectral filtering operator. Let $\mathcal{F}$ and $\mathcal{F}^{-1}$ denote the 2D Fourier transform and its inverse. Given features $Z^\ell$, FDConv performs
\begin{equation}
H^\ell = \mathrm{FDConv}(Z^\ell) = \mathcal{F}^{-1}\!\left(\mathcal{F}(Z^\ell)\odot K^\ell\right),
\label{eq:fdconv}
\end{equation}
where $\odot$ denotes element-wise multiplication in the frequency domain and $K^\ell$ is a learnable frequency response (parameterized to match the spectrum size of $\mathcal{F}(Z^\ell)$). FDConv explicitly learns a frequency response to attenuate components that are unstable across modalities.
Frequency-domain convolution effectively decouples illumination variations  from edge structures, thereby enhancing alignment robustness.
 Compared with purely spatial convolutions, the FFT-based filtering provides a cheap global receptive field and yields more stable refinement under extreme illumination.

The fused feature for scale $\ell$ is then obtained as:
\begin{equation}
M^{\ell}
=
H^{\ell}
\odot
F_{\mathrm{SVE}}^{\ell},
\qquad
F_{\mathrm{align}}^{\ell}
=
M^{\ell}
+
F_{\mathrm{evt}}^{\ell}.
\end{equation}

This coarse-to-fine refinement progressively aligns event-derived structures with SVE radiometric 
cues, yielding a final aligned feature pyramid
\begin{equation}
\mathcal{F}_{\mathrm{align}}
=
\{ F_{\mathrm{align}}^{\ell} \}_{\ell=1}^{L},
\end{equation}
which is subsequently used by the SSF fusion module to reconstruct the HDR output.

To regularize fine alignment without dense ground-truth flow, we enforce edge consistency after the coarse warp.{ We first compute an event-derived edge map $\mathrm{Edge}_{\mathrm{evt}}$ from the coarse-warped event representation. For the SVE reference, we use an exposure-normalized intensity map $I_{\mathrm{SVE}}^{\mathrm{ref}}$.} We define
\begin{equation}
L_{\mathrm{align}} = \frac{1}{HW}\sum_{\mathbf{p}}
\left\|\nabla I_{\mathrm{SVE}}^{\mathrm{ref}}(\mathbf{p}) - \mathrm{Edge}_{\mathrm{evt}}(\mathbf{p})\right\|_1,
\label{eq:alignloss}
\end{equation}
which penalizes residual geometric mismatch and stabilizes the subsequent fusion stage.

\subsubsection{Fusion and High Dynamic Range Reconstruction}
\label{subsec:fusion_hdr}

After coarse-to-fine alignment, the two modalities provide geometrically consistent but semantically complementary features.
{The SVE branch encodes exposure-diverse radiometric cues, while the event branch emphasizes high-frequency.}
We fuse them using a convolutional cross-modal interaction block, implemented by channel concatenation followed by residual $3\times 3$ convolutions and gating.

Let $\{F_{\mathrm{SVE}}^{\ell}\}_{\ell=1}^{L}$ denote the multi-exposure SVE feature pyramid obtained after demultiplexing and encoding, and
$\{F_{\mathrm{evt}}^{\text{aligned},\ell}\}_{\ell=1}^{L}$ the corresponding
aligned event features. At each encoder level $\ell$, we concatenate the
modalities along the channel dimension:
\begin{equation}
Z^{(\ell)} = 
\left[
F_{\mathrm{SVE}}^{(\ell)}
\;\Vert\;
F_{\mathrm{evt}}^{\text{aligned},(\ell)}
\right],
\end{equation}
and feed the result into a modality-fusion block implemented by a sequence of
$3 \times 3$ convolutions and nonlinear activations:
\begin{equation}
F_{\mathrm{fuse}}^{(\ell)}
=
\mathcal{C}_{\ell}\!\left(Z^{(\ell)}\right),
\end{equation}
where $\mathcal{C}_{\ell}$ denotes a learnable convolutional operator.
This block learns how to combine radiometric structure from SVE features with
motion- and edge-dominant cues from the event branch without assuming
modality-specific priors. During decoding, skip connections from each fusion
layer guide the reconstruction of the final HDR image $\hat{I}_{\mathrm{HDR}}$:
\begin{equation}
\hat{I}_{\mathrm{HDR}} = \mathrm{Dec}\!\left(\{F_{\mathrm{fuse}}^{(\ell)}\}\right).
\end{equation}

To encourage modality-invariant structure, we regularize the fused representation using
mutual information (MI). At each pyramid level $\ell$, we apply global average pooling to obtain
feature vectors $\mathbf{u}^{(\ell)}=\mathrm{GAP}(F^{(\ell)}_{\mathrm{SVE}})$ and
$\mathbf{v}^{(\ell)}=\mathrm{GAP}(F^{(\ell)}_{\mathrm{evt}})$.
Following MINE~\cite{pmlr-v80-belghazi18a}, we maximize a variational lower bound of
$I(\mathbf{u}^{(\ell)};\mathbf{v}^{(\ell)})$ parameterized by a critic network $T_{\phi}$:
\begin{equation}
\widehat{I}_{\phi}(\mathbf{u};\mathbf{v})
=
\mathbb{E}_{P_{\mathrm{joint}}}\!\big[T_{\phi}(\mathbf{u},\mathbf{v})\big]
-
\log \mathbb{E}_{P_{\mathrm{marg}}}\!\big[\exp(T_{\phi}(\mathbf{u},\mathbf{v}))\big],
\end{equation}
where $P_{\mathrm{joint}}$ samples paired features from the same time window and $P_{\mathrm{marg}}$
is formed by shuffling $\mathbf{v}$ within the mini-batch.
We use $\mathcal{L}_{\mathrm{MI}}=-\sum_{\ell}\widehat{I}_{\phi}(\mathbf{u}^{(\ell)};\mathbf{v}^{(\ell)})$
and optimize it jointly with the reconstruction losses.

Radiometric fidelity and perceptual quality are supervised via typical HDR losses:
\begin{equation}
\mathcal{L}_{\mathrm{rec}}=\|\hat{I}_{\mathrm{HDR}}-I_{\mathrm{GT}}\|_1,\qquad
\mathcal{L}_{\mathrm{SSIM}}=1-\mathrm{SSIM}(\hat{I}_{\mathrm{HDR}},I_{\mathrm{GT}}),
\end{equation}
\begin{equation}
\mathcal{L}_{\mathrm{VGG}}
=
\|
\Phi_{\mathrm{VGG}}(\hat{I}_{\mathrm{HDR}})
-
\Phi_{\mathrm{VGG}}(I_{\mathrm{GT}})
\|_1.
\end{equation}
These losses ensure radiometric correctness while preserving perceptual detail.

\subsubsection{Learnable Fusion Loss}
Although the above losses supervise HDR fidelity, they weight SVE and event cues in a fixed manner. In practice, the optimal weighting is spatially varying: SVE is reliable in non-saturated regions, while events provide stronger constraints in over-exposed and motion-blurred areas. We therefore introduce a learnable fusion loss that predicts pixel-wise modality weights. {Fixed fusion weights assume uniform reliability of SVE and event cues, which is violated in our asymmetric setup.
SVE becomes unreliable in saturated/under-exposed regions, while events can be sparse or locally inconsistent under residual parallax. We thus learn pixel-wise weights to adaptively emphasize the more reliable modality per region. Compared with fixed-weight fusion, the learnable weighting improves highlight/shadow recovery, and is more robust to local misalignment by down-weighting inconsistent constraints. The pseudocode is shown in Algorithm 1.}

A lightweight modulation network $G_\theta$ takes the fused bottleneck features and outputs two logit maps, which are converted into weights via a softmax:
\begin{equation}
[w_{\mathrm{exp}}(\mathbf{p}),\, w_{\mathrm{evt}}(\mathbf{p})] = \mathrm{softmax}\!\left(G_\theta(F_{\mathrm{SVE}},F_{\mathrm{evt}}^{\mathrm{align}})\right),\quad
w_{\mathrm{exp}}(\mathbf{p})+w_{\mathrm{evt}}(\mathbf{p})=1.
\label{eq:weight}
\end{equation}

We define a learnable intensity-consistency term
\begin{equation}
L_{\mathrm{int}}=\frac{1}{HW}\sum_{\mathbf{p}}
\Big(
w_{\mathrm{exp}}(\mathbf{p})\,\|\hat I_{\mathrm{HDR}}(\mathbf{p})-I_{\mathrm{SVE}}^{\mathrm{ref}}(\mathbf{p})\|_1
+
w_{\mathrm{evt}}(\mathbf{p})\,\|\nabla_t \hat L_{\mathrm{HDR}}(\mathbf{p})-\Delta L_{\mathrm{evt}}(\mathbf{p})\|_1
\Big),
\label{eq:lint}
\end{equation}
where $\hat L_{\mathrm{HDR}}=\log(\hat I_{\mathrm{HDR}}+\epsilon)$, $\nabla_t$ is the temporal finite difference between consecutive predictions, and $\Delta L_{\mathrm{evt}}(\mathbf{p})$ denotes the event-induced log-intensity change estimated from events accumulated within the corresponding trigger interval.

To exploit the edge sensitivity of events, we further use a gradient-consistency loss
\begin{equation}
L_{\mathrm{grad}}=\frac{1}{HW}\sum_{\mathbf{p}}
\left\|\nabla \hat I_{\mathrm{HDR}}(\mathbf{p})-\mathrm{Edge}_{\mathrm{evt}}(\mathbf{p})\right\|_1.
\label{eq:lgrad}
\end{equation}

The learnable fusion loss is
\begin{equation}
L_{\mathrm{fuse}} = L_{\mathrm{int}} + \alpha L_{\mathrm{grad}},
\label{eq:lfuse}
\end{equation}
and the total objective is
\begin{equation}
L_{\mathrm{total}}=\lambda_1 L_{\mathrm{rec}}+\lambda_2 L_{\mathrm{SSIM}}+\lambda_3 L_{\mathrm{VGG}}+\lambda_4 L_{\mathrm{MI}}+\lambda_5 L_{\mathrm{align}}+\lambda_6 L_{\mathrm{fuse}}.
\label{eq:ltotal}
\end{equation}
The parameters of $G_\theta$ are optimized jointly with the reconstruction network through back-propagation.

\begin{figure}[t]
\centering
\small
\begingroup
\setlength{\tabcolsep}{0pt}
\renewcommand{\arraystretch}{1.15}
\begin{tabular}{@{}p{0.97\linewidth}@{}}
\toprule
\textbf{Algorithm 1: Learnable fusion loss (per iteration).}\\[-2pt]
\textbf{Input:} SVE frame $M$, event representation $E$, SVE/event features $F_{\mathrm{SVE}},F_{\mathrm{evt}}$.\\
\textbf{Output:} Total loss $L_{\mathrm{total}}$ and updated parameters $(\Theta,\theta)$.\\[2pt]

\textbf{1.} $\hat{I}\leftarrow \mathrm{Net}_{\Theta}(M,E)$.\\
\textbf{2.} $(w_{\mathrm{exp}},w_{\mathrm{evt}})\leftarrow 
\mathrm{softmax}\!\big(G_{\theta}([F_{\mathrm{SVE}},F_{\mathrm{evt}}])\big)$.\\
\textbf{3.} Compute $L_{\mathrm{int}}$ and $L_{\mathrm{grad}}$.\\
\textbf{4.} $L_{\mathrm{fuse}}\leftarrow L_{\mathrm{int}}+\alpha\,L_{\mathrm{grad}}$.\\
\textbf{5.} $L_{\mathrm{total}}\leftarrow \sum_k \lambda_k L_k$ .\\
\textbf{6.} Update $(\Theta,\theta)$ jointly by backprop on $L_{\mathrm{total}}$.\\
\bottomrule
\end{tabular}
\vspace{-4pt}

\endgroup
\label{alg:fusionloss}
\end{figure}

\section{Experiments}
\label{sec:experiments}
We evaluate the proposed Event-SVE HDR reconstruction framework on both synthetic benchmarks and real captures. For synthetic data with ground truth, we report { peak signal-to-noise ratio (PSNR), structural similarity index measure (SSIM), and  learned perceptual image patch similarity (LPIPS)}. For real data without HDR ground truth, we use no-reference perceptual metrics natural image quality evaluator (NIQE)\cite{NIQ2013}, perception-based image quality evaluator (PIQE)\cite{PIQE2015},  and image entropy\cite{Entropy2021}, and we complement all quantitative results with qualitative comparisons to reveal highlight recovery, edge fidelity, and ghosting artifacts.

\subsection{Implementation Details}

Training requires triplets $(E, M, L)$ consisting of events, SVE observations, and HDR ground truth. Since no public dataset natively provides Event-SVE-HDR triplets at sufficient scale, we adopt the synthetic dataset and generation pipeline released with HDRNet/HDRev~\cite{HDRev2023learningcvpr}, which already provides paired event streams, intensity frames, and HDR ground-truth images, together with the corresponding motion and noise settings. { Concretely, events are generated using an event camera simulator~\cite{Rebecq2018ESIMAO}, and image formation follows a physically based rendering model~\cite{endoSA2017}, with realism augmentations including randomly sampled camera response exposure times ~\cite{guo2019cvpr}. Given the provided HDR ground truth, we simulate the $2\times2$ macro-pixel micro-attenuation observation by applying four attenuation factors $\{\tau_k\}$ and forming a single raw mosaic frame according to the SVE encoding model. To emulate non-coaxial discrepancies, we further apply small random 2D perturbations between the simulated SVE frames and the event-derived frames. All other components, including event generation and noise models, follow HDRNet/HDRev~\cite{HDRev2023learningcvpr} unchanged.}

For sample construction, each SVE frame is paired with a single trigger-aligned event window $T_i=[t_i,\; t_{i+1})$: we use one window per frame, no overlap, and a fixed window length.

We implement a lightweight recurrent U-Net with two downsampling layers, unrolled for $T=6$ steps. The framework is developed in PyTorch and trained on four NVIDIA GeForce RTX 4090 GPUs, with initialization from HDRev-Net~\cite{HDRev2023learningcvpr}. Loss weights $\{\lambda_k\}$ are selected on a held-out validation set: starting from the baseline reconstruction weights, we tune newly introduced terms via a small grid search while keeping gradient magnitudes across terms comparable in early epochs. {On an NVIDIA GeForce RTX 4090 GPU with batch size 1, our current PyTorch implementation processes one frame in approximately $100$\,ms for the full pipeline (coarse alignment + refinement + HDR reconstruction) at input resolutions of $1024\times1224$ (SVE) and $1280\times720$ (events), corresponding to $\sim10$ FPS, with $\sim18$\,GB peak GPU memory and $87$\,M parameters.}

\begin{figure*}[t]
  \centering
  \includegraphics[width=\linewidth]{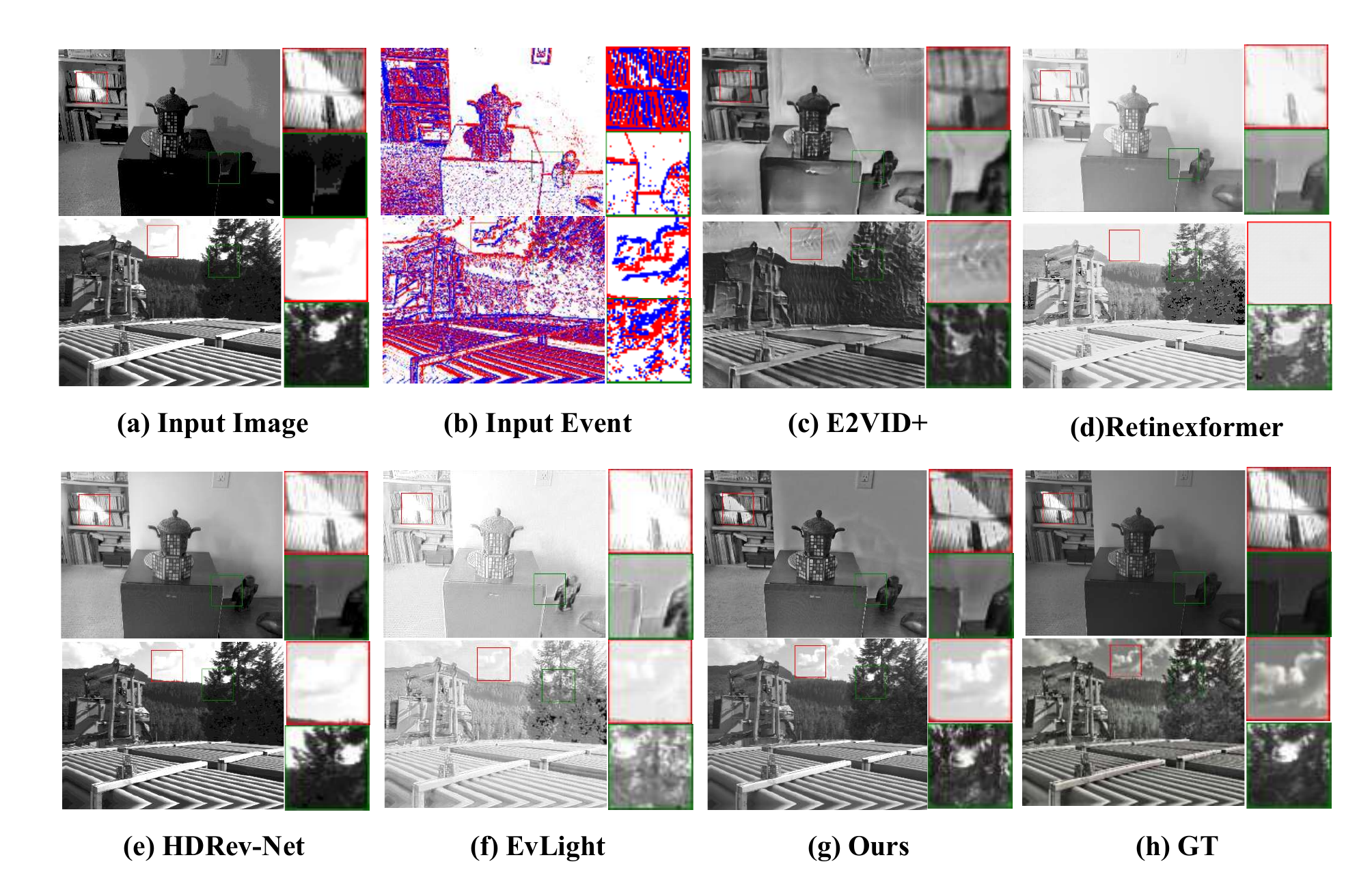}
  \caption{Qualitative comparison on the SVE-HDR dataset. Zoom in for better visualization of details. Our results provide more details and visual effects than other methods.}
  \label{fig:visual_results}
\end{figure*}

\subsection{Simulation Experiment}

To evaluate the effectiveness of our HDR fusion pipeline, we compare against state-of-the-art HDR imaging methods. For traditional methods, we follow the authors’ recommended parameter settings. For deep learning methods, we adopt the inference protocol described in the original papers, with dataset-specific adjustments to ensure fairness. The comparative studies span multiple categories, including frame-based reconstruction (Liu et al. \cite{liu2023low}), the colored event-based method E2VID \cite{rebecq2019high}, event-guided approaches like Han et al. \cite{han2020cvpr} and HDRev-Net\cite{HDRev2023learningcvpr}. None of the models were fine-tuned.

\subsubsection{Quantitative Results}

Table~\ref{synthetic} reports quantitative results on the test set. We adopt PSNR and SSIM as full-reference fidelity measures, and LPIPS as a perceptual similarity metric. 
Overall, our method achieves the best PSNR and SSIM, and the lowest LPIPS\cite{Zhang2018TheUE}, indicating improved radiometric fidelity and more accurate structural reconstruction under extreme illumination and motion. 
Compared with the strongest baseline, we consistently improve both PSNR and SSIM, suggesting that the gain comes not only from local sharpening but also from globally more stable brightness and exposure normalization. {In saturated highlights, the SVE sub-exposures with stronger attenuation preserve unsaturated measurements, while the event stream provides sharp structural cues that help avoid halo artifacts and maintain edge consistency. In deep shadows, event cues compensate for low-SNR SVE observations by emphasizing contrast transitions, improving structural fidelity even when radiometric details are weak.}

\begin{table*}[t]
\centering
\caption{{Quantitative comparison on the synthetic HDR dataset. All methods are evaluated on the same test split with synchronized event windows and SVE-like multi-exposure inputs.
We report PSNR/SSIM ($\uparrow$) and LPIPS ($\downarrow$); best and second-best are highlighted.}}
\label{synthetic}
\resizebox{0.95\textwidth}{!}{
\begin{tabular}{lcccccc}
\toprule
Method & 
E2VID~\cite{rebecq2019high} &
Liu et al.~\cite{liu2023low} &
Han et al.~\cite{han2020cvpr} &
Li et al.~\cite{li2020tip} &
HDRev-Net~\cite{HDRev2023learningcvpr} &
\textbf{Ours} \\
\midrule
PSNR $\uparrow$  & 13.734 & 23.159 & 20.697 & 20.673 & \underline{24.071} & \textbf{24.241} \\
SSIM $\uparrow$   & 0.589  & 0.901  & 0.861  & 0.890  & \underline{0.928}  & \textbf{0.935} \\
LPIPS$\downarrow$  & 0.451  & \underline{0.104}  & 0.208  & 0.151  & 0.110  & \textbf{0.102}  \\
\bottomrule
\end{tabular}
}
\end{table*}

\subsubsection{Visual Results}

Fig.~\ref{fig:visual_results} presents qualitative comparisons among different methods on challenging scenes with extreme lighting and fast motion. Frame-based HDR methods suffer from over-exposure due to the lack of precise temporal alignment in dynamic regions. Event-based methods \cite{rebecq2019high} capture sharp edges and motion details but lack rich texture and suffer from low-frequency flicker. 
The fusion-based HDRev-Net\cite{HDRev2023learningcvpr} produces better detail than single-modality methods but still exhibits residual parallax near depth discontinuities due to imperfect alignment. In contrast, {our method preserves both fine textures and motion sharpness. The proposed fine alignment stage effectively corrects residual misalignments, resulting in cleaner HDR reconstructions.}

\begin{table}[b]
\centering
\caption{No-reference quality assessment on real captures. Lower (higher) is better for NIQE/PIQE (Entropy).}
\label{tab:optica_metrics}
\resizebox{0.9\textwidth}{!}{
\begin{tabular}{lcccccc}
\toprule
Metric & SPD-MEF~\cite{spa_mef} & HDRev-Net~\cite{HDRev2023learningcvpr} & E2VID+~\cite{rebecq2019high} & EvLight~\cite{liang2024towards} & Retinexformer~\cite{cai2023retinexformer} & \textbf{Ours} \\
\midrule
NIQE$\downarrow$   & \textbf{8.33}  & 9.43  & 16.44 & 14.38  & 17.29 & 10.97 \\
PIQE$\downarrow$   & 14.18 & 22.81 & 30.06 & 16.16  & 35.67 & \textbf{12.82} \\
Entropy$\uparrow$  & 5.33 & 4.65  & 6.47 & 5.27  & 6.30 & \textbf{6.91} \\
\bottomrule
\end{tabular}
}
\end{table}

\subsection{Experiments on Real Data}

We further validate our system using the hardware prototype described in Section~\ref{subsec:hardware}. Event streams and SVE frames are synchronized through the hardware trigger described earlier.

\begin{figure}[t]
\centering
\includegraphics[width=\linewidth]{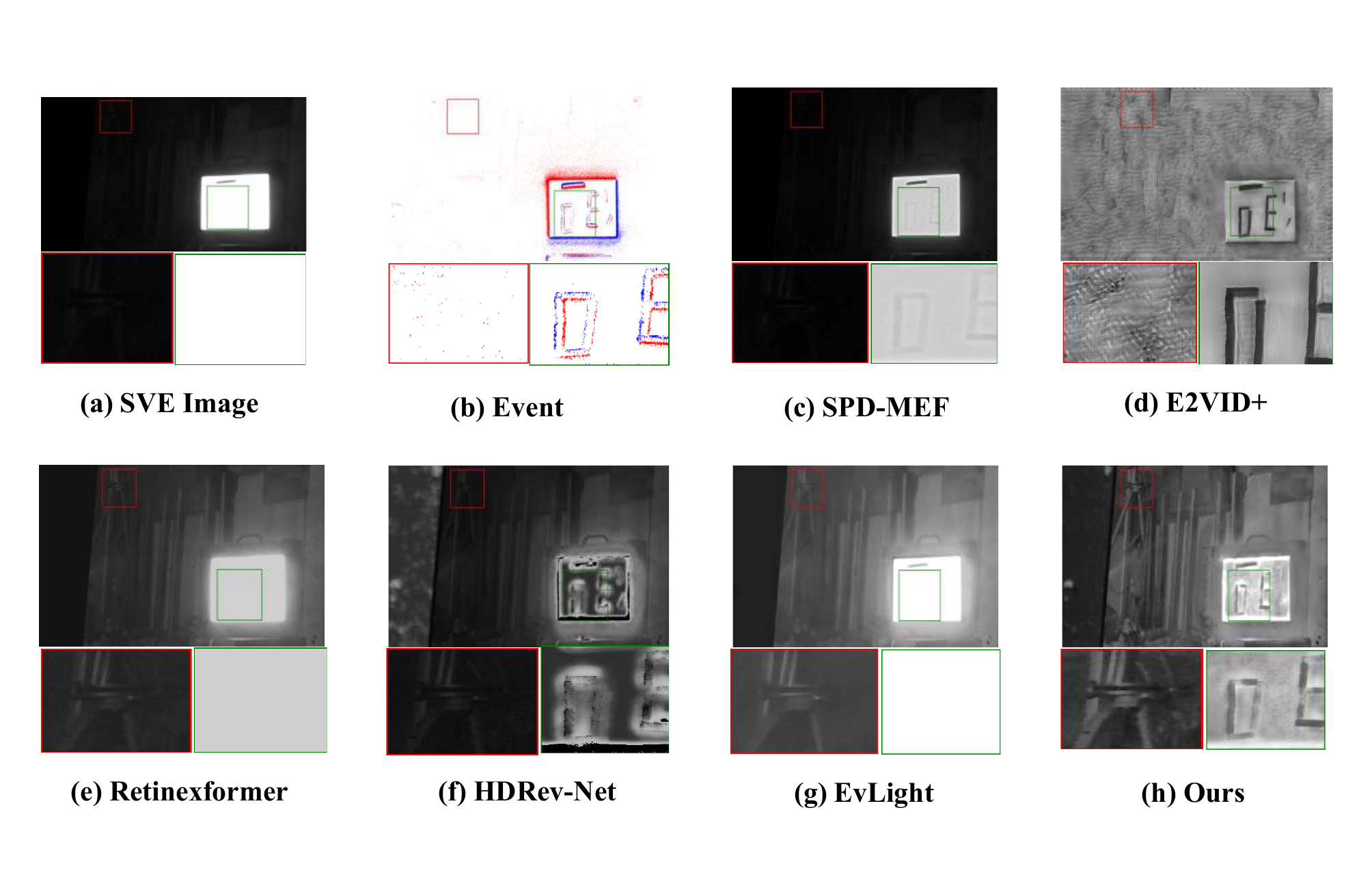}

\caption{Real-world HDR reconstruction results captured by the proposed Event-SVE rig. Compared to frame-based methods, which suffer from overexposure in bright regions, and event-based approaches, which lack background context, our method provides more comprehensive information. Moreover, unlike other event-frame fusion techniques, our approach preserves clearer textures in both overexposed and background regions, delivering more useful visual details.}

\label{fig:qualitative_real}
\end{figure}

Fig.~\ref{fig:qualitative_real} shows the reconstruction results of representative real-world scenarios captured by our hardware prototype.
Despite strong radiometric variation, our method produces visually consistent HDR results with reduced ghosting and clearer structural boundaries. We compared our methods with those based on image enhancement~\cite{cai2023retinexformer,spa_mef}, event data reconstruction~\cite{rebecq2019high}, and event image fusion~\cite{HDRev2023learningcvpr, liang2024towards}. 

To assess reconstruction quality on real captures where HDR ground truth is unavailable, we report no-reference perceptual metrics NIQE and PIQE, together with image entropy as a statistic reflecting information richness. Lower NIQE/PIQE indicates better perceptual quality, while higher entropy generally suggests richer details. As shown in Table~\ref{tab:optica_metrics}, our method achieves the best PIQE and the highest entropy, indicating improved perceptual sharpness and information richness, while NIQE is competitive but not the best. This is because compared to conventional cameras, SVE can capture more exposure parameters simultaneously, combined with the high dynamics of event cameras, thereby achieving better visual effects and more information.

\subsubsection{Qualitative Results}

To complement the quantitative evaluation, we provide extensive qualitative comparisons across diverse scenes, including high-contrast indoor environments, outdoor nighttime scenarios, and sequences containing rapid motion or extreme illumination variation. As shown in Fig.~\ref{fig:qualitative_real}, conventional HDR reconstruction methods often suffer from residual noise amplification, local overexposure, or the loss of fine spatial details, especially in highlight-saturated regions or dark areas with limited photon counts.
In contrast, our method consistently produces visually more faithful results with better global illumination balance and sharper structural boundaries. The proposed alignment-aware reconstruction effectively suppresses ghosting artifacts in dynamic regions and preserves local textures such as edges, thin structures, and high-frequency patterns. Moreover, {dark regions are restored with richer gradients, while bright regions remain well-preserved with natural highlight roll-off. Since no-reference image-quality metrics are not entirely consistent with human perception for HDR tone mapping and cross-modal alignment, we complement them with qualitative comparisons, allowing for direct evaluation of ghosting suppression, edge fidelity, and highlight recovery.}
Overall, the qualitative visualizations demonstrate that our approach achieves a more natural appearance, superior local contrast, and stronger robustness across challenging illumination conditions, corroborating the advantages observed in the quantitative metrics.

\subsection{Ablation Studies}

The ablation results in Table~\ref{tab:ablation_nr} systematically validate the necessity of each component in our cross-modal HDR pipeline.
Removing the coarse alignment module leads to the largest degradation in NIQE and PIQE due to pronounced parallax and structural inconsistencies between the SVE frames and event edges. Eliminating the fine alignment stage further introduces ghosting artifacts, resulting in elevated perceptual distortion and reduced image entropy. Disabling the FDConv block produces blurred boundaries and weak gradient responses, which are reflected in both higher PIQE and lower entropy. Finally, replacing the learnable fusion loss with fixed weighting reduces the network’s ability to adaptively balance radiometric and temporal cues, causing notable declines in highlight recovery and global contrast.
Across the ablations, the full system achieves the best overall trade-off. It yields the strongest PIQE and entropy improvements and maintains comparable NIQE, demonstrating that accurate cross-modal alignment and adaptive fusion are jointly important for robust HDR reconstruction in the asymmetric Event-SVE setting.

\begin{table}[t]
\centering
\caption{Ablation study on major system components using no-reference metrics. 
Arrows indicate whether lower ($\downarrow$) or higher ($\uparrow$) is better.}
\label{tab:ablation_nr}
\resizebox{0.85\linewidth}{!}{
\begin{tabular}{lccc c}
\toprule
Removed Component 
& NIQE$\downarrow$  & PIQE$\downarrow$ & Entropy$\uparrow$ & Notes \\ 
\midrule
No alignment         
& 13.81 & 34.24 & 5.52 & Blur ghosting \\
No learnable fusion loss    
& 10.94 & 26.32 & 6.31 & reduced highlight recovery \\
\textbf{Full system (Ours)} 
& \textbf{10.91} & \textbf{25.80} & \textbf{6.91} & best  \\
\bottomrule
\end{tabular}
}
\end{table}

\subsection{Discussion}

{The results demonstrate that the proposed asymmetric Event-SVE system successfully overcomes the geometric and radiometric challenges typically associated with asymmetric dual-modality HDR image acquisition.} The proposed coarse-to-fine alignment and learnable fusion mechanisms substantially improve structural fidelity and highlight preservation, particularly in scenes exhibiting extreme illumination contrast or rapid motion where single-modality solutions fail. Real-world experiments further confirm that integrating SVE radiometric cues with event-driven temporal precision yields robust HDR reconstruction across diverse conditions, highlighting the practical value of our joint design in high-dynamic-range environments. 

{Although the proposed Event-SVE fusion framework performs robustly in most tested scenarios, we observe corner cases where the reconstruction quality degrades. In particular, extremely fast motion may cause severe blur in the SVE frame, which weakens cross-modal correspondences for coarse alignment and reduces the reliability of radiometric cues. Moreover, in nighttime or very low-light scenes, the SVE measurements can approach the noise floor, while the event stream may become sparse or contaminated by background activity, diminishing the effectiveness of event-driven structural constraints and leading to unclear outputs. These limitations suggest future improvements such as motion-aware deblurring, event reliability gating and denoising, and acquisition tuning for low-light conditions.}

\section{Conclusion}
\label{Conclusion}

We presented an asymmetric{ hybrid } HDR imaging system that unifies a SVE camera and an event camera through a principled hardware-algorithm co-design. The heterogeneous optical paths and non-coaxial geometry of the physical setup necessitated a dedicated coarse-to-fine alignment strategy that combines calibration-driven rectification with refinement via spatial pooling and frequency-domain filtering, effectively compensating for non-common-path distortions and motion-induced inconsistencies typical of hybrid imaging configurations. On this aligned foundation, we developed a cross-modal HDR reconstruction framework that {jointly exploits the radiometric completeness of SVE mosaics, high-contrast responsiveness of event streams}. Convolutional fusion, mutual-information regularization, and a learnable fusion loss enable the system to dynamically reweight modality contributions, yielding stable performance under {extreme illumination and large exposure disparities}. Experiments on synthetic benchmarks and real captures demonstrate consistent gains in highlight recovery, edge fidelity, and temporal robustness over existing HDR techniques. Future work includes tighter optical-computational integration, end-to-end cross-modal calibration, adaptive event-threshold modeling, temporally coherent HDR video reconstruction, and real-time deployment for robotics, autonomous perception, and high-speed scientific imaging.

\section*{Acknowledgement}
This work was supported by the National Natural Science Foundation of China (Grant No. 12372189).

\section*{Disclosures}
The authors have no conflicts to disclose.

\section*{Data availability}
The data that support the findings of this study are available from the corresponding authors upon reasonable request.

\end{document}